\documentclass[letterpaper]{article} % DO NOT CHANGE THIS
\usepackage{aaai2026}  % DO NOT CHANGE THIS
\usepackage{times}  % DO NOT CHANGE THIS
\usepackage{helvet}  % DO NOT CHANGE THIS
\usepackage{courier}  % DO NOT CHANGE THIS
\usepackage[hyphens]{url}  % DO NOT CHANGE THIS
\usepackage{graphicx} % DO NOT CHANGE THIS
\usepackage{natbib}  % DO NOT CHANGE THIS AND DO NOT ADD ANY OPTIONS TO IT
\usepackage{caption} % DO NOT CHANGE THIS AND DO NOT ADD ANY OPTIONS TO IT
\usepackage{algorithm}
\usepackage{algorithmic}

\usepackage{multirow}
\usepackage{booktabs}
\usepackage{pifont}
\usepackage{arydshln}
\usepackage{amsmath}
\usepackage{amssymb}

\usepackage{newfloat}
\usepackage{listings}
\DeclareCaptionStyle{ruled}{labelfont=normalfont,labelsep=colon,strut=off} % DO NOT CHANGE THIS
\floatstyle{ruled}
\newfloat{listing}{tb}{lst}{}
\floatname{listing}{Listing}

\title{Beyond Quadratic: Linear-Time Change Detection with RWKV}
\author {
    Zhenyu Yang\textsuperscript{\rm 1,\rm 2},
    Gensheng Pei\textsuperscript{\rm 3}, 
    Tao Chen\textsuperscript{\rm 1,\rm 2}, 
    Xia Yuan\textsuperscript{\rm 1},\\
    Haofeng Zhang\textsuperscript{\rm 1}, 
    Xiangbo Shu\textsuperscript{\rm 1}, 
    Yazhou Yao\textsuperscript{\rm 1,\rm 2}\thanks{Corresponding Author}
}
\affiliations {
    \textsuperscript{\rm 1}School of Computer Science and Engineering, Nanjing University of Science and Technology, Nanjing, China\\
    \textsuperscript{\rm 2}State Key Laboratory of Intelligent Manufacturing of Advanced Construction Machinery, China\\
    \textsuperscript{\rm 3}Department of Electrical and Computer Engineering, Sungkyunkwan University, Suwon, Korea\\
    zhenyu\_yang@njust.edu.cn, peigsh@skku.edu, \{taochen, yuanxia, zhanghf, shuxb, yazhou.yao\}@njust.edu.cn
}

\begin{document}

\maketitle

\begin{abstract}
Existing paradigms for remote sensing change detection are caught in a trade-off: CNNs excel at efficiency but lack global context, while Transformers capture long-range dependencies at a prohibitive computational cost. This paper introduces \textbf{ChangeRWKV}, a new architecture that reconciles this conflict. By building upon the Receptance Weighted Key Value (RWKV) framework, our ChangeRWKV uniquely combines the parallelizable training of Transformers with the \textit{linear-time} inference of RNNs.
Our approach core features two key innovations: a \textit{hierarchical} RWKV encoder that builds multi-resolution feature representation, and a novel \textit{Spatial-Temporal} Fusion Module (STFM) engineered to resolve \textit{spatial} misalignments across scales while distilling fine-grained \textit{temporal} discrepancies.
ChangeRWKV not only achieves state-of-the-art performance on the LEVIR-CD benchmark, with an \textbf{85.46\% IoU} and \textbf{92.16\% F1} score, but does so while drastically reducing parameters and FLOPs compared to previous leading methods. This work demonstrates a new, efficient, and powerful paradigm for operational-scale change detection. Our code and model are publicly available.
\end{abstract}

\begin{links}
\link{Code}{https://github.com/ChangeRWKV/ChangeRWKV}
\end{links}

\section{Introduction}
Remote sensing change detection (RSCD), the task of identifying meaningful differences from multi-temporal imagery, is fundamental to applications like environmental monitoring, urban planning, and disaster assessment. A fundamental tension defines the recent evolution of RSCD architectures: while high-resolution imagery demands powerful models capable of capturing long-range spatial context, critical applications such as rapid post-disaster analysis on unmanned aerial vehicles (UAVs) require lightweight, low-latency models for on-device deployment.

\begin{figure}[t]
\centering
\includegraphics[width=1.0\linewidth]{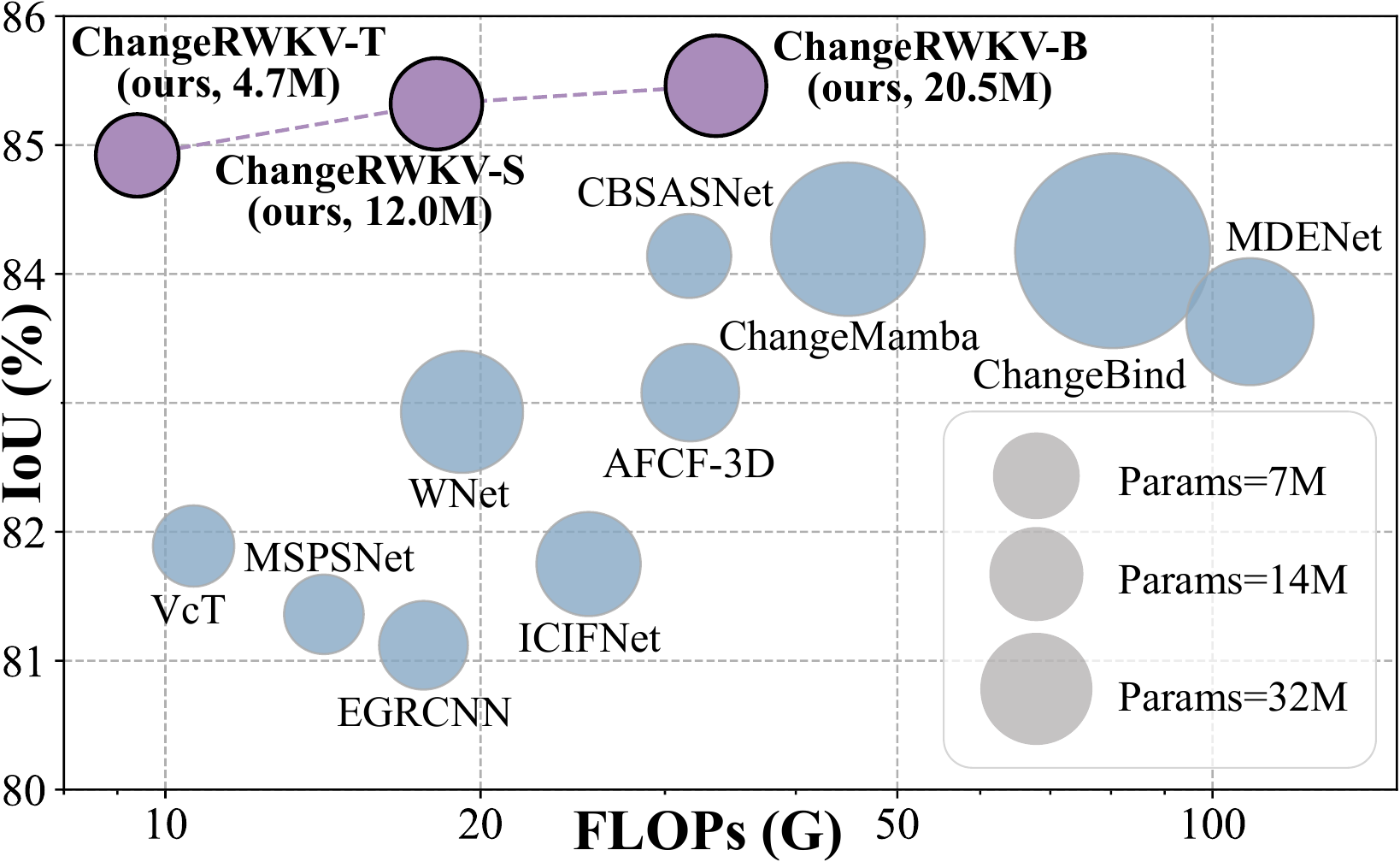}
\caption{Efficiency versus accuracy on the LEVIR-CD benchmark. The proposed ChangeRWKV family establishes a new state-of-the-art frontier, delivering superior IoU scores while demanding significantly fewer computational resources (FLOPs) and parameters than existing methods. Our tiny model, for instance, achieves a competitive\textbf{ 84.92\%} IoU with only \textbf{4.7M} Params and \textbf{9.40G} FLOPs.}
\label{fig:blob}
\end{figure}

Early approaches \cite{zheng2021change, ye2023adjacent, huang2024spatiotemporal, pei2024videomac, cai2024poly, pei2022hierarchical} based on convolutional neural networks (CNNs) are computationally efficient and excel at extracting local features. However, their inherently local receptive fields limit their ability to model the global context essential for disambiguating complex changes. Vision Transformers (ViTs) \cite{dosovitskiy2021vit, liu2021swin, xie2021segformer, bernhard2023mapformer, chen2024multi, zhou2025unialign} overcame this limitation by using self-attention to model global relationships, leading to significant accuracy gains \cite{chen2024multi,zang2025changediff, benidir2025change}. This performance, however, comes at the cost of quadratic complexity, rendering standard ViTs impractical for the high-resolution images common in remote sensing. This bottleneck has spurred research into linear-time architectures, such as state-space models like Mamba \cite{mamba, mamba2}, which promise both global modeling and linear scalability.

Recently, the Receptance Weighted Key Value (RWKV) architecture \cite{peng2023rwkv, peng2024eagle, peng2025rwkv} has emerged as a compelling alternative, blending the parallelizable training of Transformers with the linear complexity of RNNs. Unlike self-attention with its $\mathcal{O}(T^2d)$ complexity, RWKV operates with $\mathcal{O}(Td)$ complexity in both computation and memory, making it highly scalable. Its design, which decouples spatial and channel mixing, has demonstrated competitive performance and enhanced training stability over other linear-time models. While RWKV's potential has been realized in natural language and, more recently, in general vision tasks \cite{duan2024vision, zhou2024bsbprwkv, jiang2025rwkv, lv2025scalematch}, its aptitude for the nuanced task of bi-temporal image analysis in RSCD remains unexplored.

In this work, we bridge this gap by presenting Change-RWKV, a robust and efficient framework for remote sensing change detection. Our core idea is to leverage the linear complexity and strong representation power of RWKV to build a powerful yet lightweight model. We design a hierarchical encoder that processes each image to extract rich, multi-scale features with minimal computational overhead. At the heart of our model is a novel Spatial-Temporal Fusion Module (STFM), specifically engineered to effectively integrate these multi-scale features and model the temporal discrepancies between the image pairs. As shown in Fig.~\ref{fig:blob}, Change-RWKV sets a new standard for the trade-off between accuracy and efficiency, making high-performance change detection feasible even under constrained computational budgets.
Our main contributions are summarized as follows:
\begin{itemize}
    \item We propose ChangeRWKV, the \textit{first} framework to successfully adapt the RWKV architecture for remote sensing change detection, establishing a new benchmark for highly \textit{efficient} yet \textit{accurate} models.
    \item We introduce a novel STFM that adeptly integrates hierarchical features and models bi-temporal differences, significantly enhancing the model's ability to discriminate subtle and complex changes.
    \item We validate our approach through extensive experiments on four diverse optical and SAR change detection benchmarks. Our results show that ChangeRWKV achieves state-of-the-art performance while drastically reducing computational costs, confirming its suitability for real-time and resource-limited deployment scenarios.
\end{itemize}

\section{Related Work}

\noindent\textbf{CNN-based Change Detection.}
CNNs form the foundation of deep learning-based RSCD. Early methods primarily utilize fully convolutional Siamese architectures to extract and compare pixel-wise features from bi-temporal images \cite{daudt2018fully}. To improve performance, subsequent works introduce more sophisticated designs, such as nested U-Nets for enhanced multi-scale feature preservation \cite{fang2021snunet}, dual-decoders to separate change and semantic information \cite{chen2022fccdn}, and specialized modules for edge-aware fusion or 3D spatio-temporal modeling \cite{huang2024spatiotemporal, ye2023adjacent}. While computationally efficient, the performance of CNN-based models is inherently constrained by their local receptive fields, which struggle to capture the global context necessary for understanding complex semantic changes. Our approach overcomes this by employing a backbone capable of long-range dependency modeling.

\noindent\textbf{Transformer-based Change Detection.}
To address the context limitations of CNNs, ViTs are adapted for RSCD. Seminal works like BIT \cite{chen2021remote} and ChangeFormer \cite{bandara2022transformer} demonstrate the power of self-attention in modeling global spatial-temporal relationships, leading to significant accuracy improvements. This paradigm is further refined through hierarchical designs like SwinSUNet \cite{zhang2022swinsunet} and hybrid models that combine convolutional priors with attention mechanisms \cite{pei2022feature, yang2025convformer}. Although Transformers excel at global context modeling, their quadratic complexity in self-attention creates a severe computational bottleneck, limiting their practical use on high-resolution imagery and in resource-constrained settings  \cite{mao2025prune}. Our work directly targets this efficiency gap without sacrificing the ability to model long-range interactions.

\noindent\textbf{Linear-Time Change Detection.}
Most recently, the field has gravitated towards linear-time architectures to balance modeling power and efficiency. This trend is dominated by methods adapting State Space Models (SSMs), particularly Mamba \cite{mamba}, for the change detection task. Architectures like ChangeMamba \cite{chen2024changemamba}, CDMamba \cite{zhang2025cdmamba}, and RS-Mamba \cite{zhao2024rs} successfully integrate Mamba's selective scan mechanism for efficient sequential modeling of image patches. Other variants explore frequency-domain analysis \cite{xing2025frequency} or change-guided feature selection \cite{ghazaei2025change} within the SSM framework. These Mamba-based models confirm the viability of linear-time architectures for RSCD, yet the exploration of this design space remains narrow. Our work diverges by being the first to investigate the RWKV architecture \cite{peng2023rwkv}, whose linear complexity, training stability, and architectural simplicity present a novel and powerful alternative for highly efficient remote sensing change detection.

\begin{figure*}[t]
    \centering
    \includegraphics[width=1\linewidth]{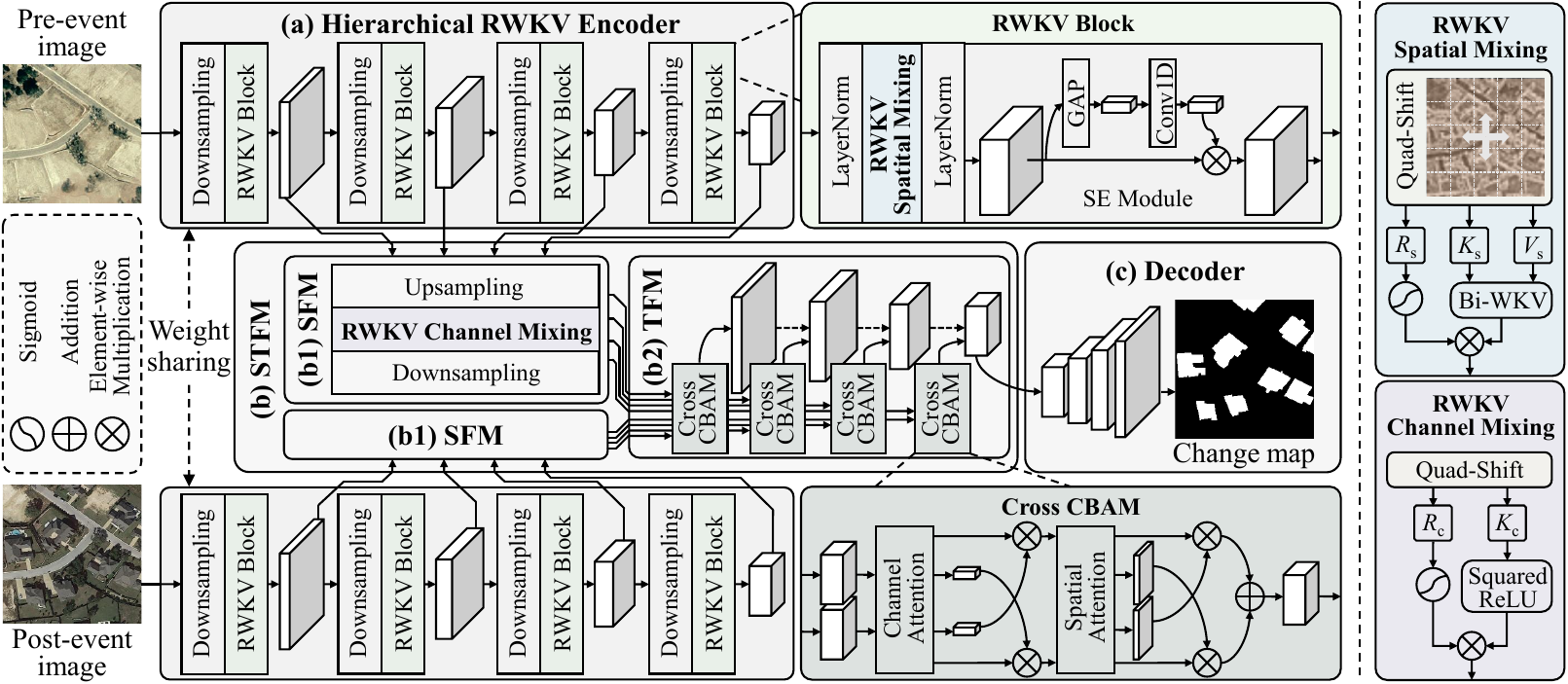}
    \caption{Overall architecture of ChangeRWKV. The model consists of three main components: (\textbf{a)} a Hierarchical RWKV Encoder that extracts multi-scale features, \textbf{(b)} a Spatial-Temporal Fusion Module (STFM) that integrates spatial and temporal cues, and \textbf{(c)} a lightweight Decoder that generates the change mask. The STFM is further decomposed into \textbf{(b1)} a Spatial Fusion Module (SFM) for multi-scale spatial alignment and \textbf{(b2)} a Temporal Fusion Module (TFM) for bi-temporal interaction.}
    \label{fig:changerwkv}
\end{figure*}

\section{Method}
\subsection{The RWKV Architecture}
The Receptance Weighted Key Value (RWKV) model \cite{peng2023rwkv, peng2024eagle, peng2025rwkv} is a novel sequence architecture that marries the parallelizable training of Transformers with the linear inference cost of RNNs. It reformulates the self-attention mechanism, which has a complexity of $\mathcal{O}(T^2d)$, into a recurrent structure with a linear complexity of $\mathcal{O}(Td)$ \textit{w.r.t.} sequence length $T$, and model dimension $d$.

At its core, RWKV processes a sequence of tokens $\{x_t\}_{t=1}^T$. For each token $x_t$, it first projects it into receptance ($r_t$), key ($k_t$), and value ($v_t$) vectors. This projection cleverly incorporates information from the previous token $x_{t-1}$ through a learnable interpolation:
\begin{equation}
    (*)_t = W_{(*)} \cdot \left[ \mu_{(*)} \odot x_t + (1 - \mu_{(*)}) \odot x_{t-1} \right],
\end{equation}
where $(*) \in \{r, k, v\}$, and $\odot$ denotes Hadamard product. $W_{(*)} \in \mathbb{R}^{d \times d}$ are the projection matrices, and $\mu_{(*)} \in \mathbb{R}^d$ are the learnable mixing coefficients.
The central mechanism is the \texttt{WKV} operation, which acts as a time-decaying, channel-wise aggregation of past values, modulated by keys:
\begin{equation}
    \texttt{WKV}_t = \frac{\sum_{i=1}^{t-1} e^{-(t-1-i)w+k_i} \odot v_i + e^{u+k_t} \odot v_t}{\sum_{i=1}^{t-1} e^{-(t-1-i)w+k_i} + e^{u+k_t}},
\end{equation}
where $w \in \mathbb{R}^d$ is a learnable decay vector that controls how much past information is forgotten, and $u \in \mathbb{R}^d$ is a learnable parameter that grounds the current token's importance. The RWKV block consists of two main sub-layers:

\noindent\textbf{Time-Mixing} models sequential dependencies. The output is gated by the receptance $r_t$, allowing the model to dynamically control how much historical context to incorporate:
\begin{equation}
    o_t = W_o \cdot (\sigma(r_t) \odot \texttt{WKV}_t),
\end{equation}
where $W_o \in \mathbb{R}^{d \times d}$ is an output projection matrix, and $\sigma(\cdot)$ is the \texttt{Sigmoid} function for modulating historical context.

\noindent\textbf{Channel-Mixing} is applied per-token to model feature interactions within each token's embedding. It is typically implemented as a simple two-layer \texttt{MLP} with a squared \texttt{ReLU} activation for enhanced non-linearity. This decoupled design of time and channel mixing contributes to RWKV's training stability and strong performance.

\subsection{Overall Architecture of ChangeRWKV}
Given a pair of co-registered remote sensing images $I_A, I_B \in \mathbb{R}^{H \times W \times C}$ captured at times $t_A$ and $t_B$, the objective of RSCD is to predict a binary change map $M \in \{0, 1\}^{H \times W}$, where $M_{m,n}=1$ if a change occurred at pixel $(m,n)$ and $0$ otherwise.
As shown in Fig. \ref{fig:changerwkv}, our Change-RWKV follows a Siamese encoder-decoder structure.

\noindent\textbf{Hierarchical RWKV Encoder:} A shared RWKV-based vision encoder processes $I_A$ and $I_B$ independently to extract hierarchical feature maps $\{f_{i1}, f_{i2}, f_{i3}, f_{i4}\}$ for each image $i \in \{A, B\}$ at four different scales.

\noindent\textbf{Spatial-Temporal Fusion Module:} In this work, we introduce a novel STFM, which processes the input bi-temporal, multi-scale features by first performing intra-image spatial fusion and then inter-image temporal fusion, yielding a set of change-centric feature maps $\{\tilde{f}_1, \tilde{f}_2, \tilde{f}_3, \tilde{f}_4\}$.

\noindent\textbf{Lightweight Decoder:} A U-Net \cite{ronneberger2015u} style decoder with skip connections takes the fused features $\{\tilde{f}_1, \tilde{f}_2, \tilde{f}_3, \tilde{f}_4\}$ and progressively upsamples them to generate the final change map $\hat{M}$.

\subsection{Hierarchical RWKV Encoder}
We adapt the sequential RWKV block for 2D vision tasks. Following the principles of VisionRWKV \cite{duan2024vision}, we replace the unidirectional time-mixing with a bidirectional spatial-mixing mechanism that aggregates information across the 2D plane, enhancing local context modeling. For efficiency, the standard channel-mixing \texttt{MLP} is replaced by a lightweight Squeeze-and-Excitation (SE) module \cite{hu2018squeeze}. Crucially, our encoder produces \textit{hierarchical} outputs, providing the \textit{multi-scale} representations essential for detecting changes of varying sizes, a key departure from flat-feature vision models.

\subsection{Spatial-Temporal Fusion Module}
The STFM is the core of our method, designed to robustly integrate features across both space and time (see Fig. \ref{fig:changerwkv}(b)).

\noindent\textbf{Spatial Fusion Module (SFM):} This module enriches features within each temporal snapshot by promoting cross-scale communication. For each image $i \in \{A, B\}$, its feature maps $\{f_{ij}\}_{j=1}^4$ are all upsampled to the finest resolution level ($H/2 \times W/2$) and concatenated along the channel dimension into a single tensor $F_i$. A residual channel-mixing block then refines this tensor:
\begin{equation}
    \hat{F_i} = F_i + \texttt{Channel-Mixing}(F_i).
\end{equation}
The refined tensor $\hat{F_i}$ is subsequently split and downsampled back to the original four scales, yielding spatially-enhanced features $\{\tilde{f}_{ij}\}_{j=1}^4$ for each image $i \in \{A, B\}$.

\begin{table*}[t]
\small
\centering
\setlength{\tabcolsep}{8.5pt}
\begin{tabular}{lcc|cccc|cccc}
\specialrule{1pt}{0pt}{0pt}
\multirow{2}{*}{\textbf{Method}} & 
\multirow{2}{*}{\textbf{\begin{tabular}[c]{@{}c@{}}Params\\ (M)\end{tabular}}} & 
\multirow{2}{*}{\textbf{\begin{tabular}[c]{@{}c@{}}FLOPs\\ (G)\end{tabular}}} & 
\multicolumn{4}{c|}{\textbf{LEVIR-CD}} & 
\multicolumn{4}{c}{\textbf{WHU-CD}} \\ % \cline{4-11}
 & & & 
\textbf{IoU} & \textbf{F1} & \textbf{P} & \textbf{R} & 
\textbf{IoU} & \textbf{F1} & \textbf{P} & \textbf{R} \\
\hline
FC-Siam-Diff \shortcite{daudt2018fully}          & 4.38  & 1.35   & 75.92 & 86.31 & 89.53 & 83.31 & 41.66 & 58.81 & 47.33 & 77.66 \\
FC-Siam-Conc \shortcite{daudt2018fully}          & 4.99  & 1.55   & 71.96 & 83.69 & 91.99 & 76.77 & 49.95 & 66.63 & 60.88 & 73.58 \\
SNUNet \shortcite{fang2021snunet}                & 12.03 & 27.44  & 78.83 & 88.16 & 89.18 & 87.17 & 71.67 & 83.50 & 85.60 & 81.49 \\
BIT \shortcite{chen2021remote}                   & 3.55  & 4.35   & 80.68 & 89.31 & 89.24 & 89.37 & 72.39 & 83.98 & 86.64 & 81.48 \\
ICIFNet \shortcite{feng2022icif}                 & 23.82 & 25.36  & 81.75 & 89.96 & 91.32 & 88.64 & 79.24 & 88.32 & 92.98 & 85.56 \\
ChangeFormer \shortcite{bandara2022transformer}  & 41.02 & 202.80 & 82.48 & 90.40 & 92.05 & 88.80 & 71.91 & 83.66 & 85.49 & 81.90 \\
WNet \shortcite{tang2023wnet}                    & 43.07 & 19.20  & 82.93 & 90.67 & 91.16 & 90.18 & 83.91 & 91.25 & 92.37 & 90.15 \\
AFCF-3D \shortcite{ye2023adjacent}               & 17.54 & 31.72  & 83.08 & 90.76 & 91.35 & 90.17 & 87.93 & 93.58 & 93.47 & 93.69 \\
CCLNet++ \shortcite{song2023toward}              & 28.78 & 23.27  & 83.62 & 91.08 & 92.31 & 89.88 & 82.32 & 90.31 & 89.83 & 90.78 \\
CF-GCN \shortcite{wang2024cf}                    & 13.58 & 43.93  & 83.41 & 90.96 & 91.75 & 90.18 & 84.90 & 91.83 & 94.81 & 89.04 \\
SEIFNet \shortcite{huang2024spatiotemporal}      & 8.37  & 27.91  & 83.40 & 90.95 & 92.49 & 89.46 & 76.04 & 86.39 & 87.01 & 85.77 \\
BiFA \shortcite{zhang2024bifa}                   & 5.58  & 53.00  & 82.96 & 90.69 & 91.52 & 89.86 & 89.34 & 94.37 & 95.15 & \underline{93.60} \\
CBSASNet \shortcite{he2024cbsasnet}              & 5.76  & 31.64  & 84.14 & 91.39 & 92.47 & 90.33 & 86.08 & 92.52 & 93.93 & 91.15 \\
ChangeBind \shortcite{noman2024changebind}       & 153.70 & 80.29 & 84.18 & 91.41 & \textbf{94.93} & 88.14 & 85.72 & 92.31 & 94.81 & 89.94 \\
ChangeMamba \shortcite{chen2024changemamba}      & 84.70 & 44.86 & 84.27 & 91.37 & \underline{93.87} & 89.00 & 88.02 & 93.63 & 94.22 & 93.05 \\
ConvFormer-CD/48 \shortcite{yang2025convformer}  & 37.72 & 5.14   & 84.23 & 91.44 & 92.37 & 90.52 & 85.41 & 92.13 & 95.14 & 89.26 \\
SPMNet \shortcite{wang2025spmnet}                & 17.33 & 6.94   & 84.07 & 90.99 & 92.12 & 90.58 & 84.84 & 91.80 & 94.67 & 89.10 \\ 
SFEARNet \shortcite{li2025sfearnet}              & 5.56  & 4.65   & 83.23 & 90.85 & 91.43 & 90.27 & 85.81 & 92.36 & 94.25 & 90.55 \\
STRobustNet \shortcite{zhang2025strobustnet}     & 5.23  & 13.19  & 83.66 & 91.11 & 91.54 & 90.67 & 83.29 & 90.89 & 92.92 & 88.94 \\
AMDANet \shortcite{su2025amdanet}                & 25.83 & 77.23  & 82.34 & 90.32 & 92.45 & 88.28 & 80.68 & 89.30 & 88.77 & 89.84 \\
\hline
\textbf{ChangeRWKV-T (ours)}                     & 4.66  & 9.40   & 84.92 & 91.85 & 92.41 & \underline{91.29} & 88.19 & 93.72 & \textbf{96.27} & 91.30 \\ 
\textbf{ChangeRWKV-S (ours)}                     & 12.00 & 18.15  & \underline{85.32} & \underline{92.08} & 93.00 & 91.17 & \textbf{90.06} & \textbf{94.77} & 95.50 & \textbf{94.05} \\ 
\textbf{ChangeRWKV-B (ours)}                     & 20.50 & 33.56  & \textbf{85.46} & \textbf{92.16} & 92.86 & \textbf{91.46} & \underline{89.59} & \underline{94.51} & \underline{96.14} & 92.93 \\ 
\specialrule{1pt}{0pt}{0pt}
\end{tabular}
\caption{Quantitative comparison on the LEVIR-CD \cite{chen2020spatial} and WHU-CD \cite{hong2023multi} test sets. All metrics are reported as a percentage (\%), with the highest value highlighted in \textbf{bold}, and the second highest \underline{underlined}.}
\label{tab:levir_whu}
\end{table*}

\noindent\textbf{Temporal Fusion Module (TFM):} After spatial enhancement, the TFM integrates the bi-temporal features at each scale $j$. Inspired by Convolutional Block Attention Module (CBAM) \cite{woo2018cbam}, we employ a new cross-attention strategy, termed as Cross CBAM (see Fig. \ref{fig:changerwkv}), to explicitly model changes. First, channel attention weights are computed from one temporal feature and applied to the other, and vice-versa, to highlight discriminative channels.
\begin{equation}
    \gamma_{ij}^{c} = \sigma(\texttt{MLP}(\texttt{GAP}(\tilde{f}_{ij}))),
\end{equation}
where \texttt{GAP} denotes Global Average Pooling. The features are then cross-refined: $\tilde{f}_{Aj}' = \gamma_{Bj}^{c} \odot \tilde{f}_{Aj}$ and $\tilde{f}_{Bj}' = \gamma_{Aj}^{c} \odot \tilde{f}_{Bj}$. Next, spatial attention maps are computed from these channel-refined features and are also cross-applied to focus on salient spatial regions of change:
\begin{equation}
    s_{ij} = \sigma(\texttt{Conv}([\texttt{AvgPool}(\tilde{f}_{ij}'); \texttt{MaxPool}(\tilde{f}_{ij}')])).
\end{equation}
The final fused feature at each level $j$ is obtained by element-wise summation of the fully refined temporal features: $\tilde{f}_j = (\tilde{f}_{Aj}' \odot s_{Bj}) + (\tilde{f}_{Bj}' \odot s_{Aj})$. This adaptive, data-driven fusion contrasts with methods relying on simple subtraction \cite{daudt2018fully, corley2024change} or predefined metrics \cite{wang2022ssa, dong2024changeclip}, allowing our model to learn optimal fusion strategies for diverse types of change.

\subsection{Loss Function}
To effectively train ChangeRWKV, we employ a hybrid loss function that combines Binary Cross Entropy (BCE) and Dice loss. The BCE loss ensures pixel-level accuracy:
\begin{equation}
    \mathcal{L}_{\text{BCE}} = -\frac{1}{N} \sum_{n=1}^{N} [ M_n \log(\hat{M}_n) + (1 - M_n) \log(1 - \hat{M}_n) ],
\end{equation}
where $N=H \times W$, $M$ is the ground-truth map, and $\hat{M}$ is the prediction. To address class imbalance and improve the segmentation of change region boundaries, we add the Dice loss, which maximizes the spatial overlap:
\begin{equation}
    \mathcal{L}_{\text{Dice}} = 1 - \frac{2 \sum M_n \hat{M}_n + \epsilon}{\sum M_n + \sum \hat{M}_n + \epsilon},
\end{equation}
where $\epsilon$ is a small constant for numerical stability. The final training objective is a weighted sum of these two losses:
\begin{equation}
    \mathcal{L} = \mathcal{L}_{\text{BCE}} + \lambda \mathcal{L}_{\text{Dice}},
\end{equation}
where $\lambda$ is a hyperparameter balancing the two terms, this composite loss promotes both high pixel-wise fidelity and structurally coherent change maps.

\section{Experiments}

\subsection{Experimental Setup}
\noindent\textbf{Datasets}.
We conduct extensive evaluations on four public benchmarks spanning both optical and synthetic aperture radar (SAR) modalities to demonstrate the effectiveness and generalization of our method. Details are as follows:

\begin{itemize}
\item \textbf{LEVIR-CD} \cite{chen2020spatial}: A widely-used dataset of 637 high-resolution (0.5m/pixel) image pairs focused on building changes in urban environments.
\item \textbf{WHU-CD} \cite{hong2023multi}: Comprises two large aerial images (0.3m/pixel) capturing significant structural evolution in Christchurch, New Zealand.
\item \textbf{LEVIR-CD+} \cite{shen2021s2looking}: An extension of LEVIR with 985 image pairs featuring longer time spans (5-14 years), and more diverse, complex change patterns.
\item \textbf{SAR-CD} \cite{alatalo2023improved}: A challenging dataset of 10,000 $512 \times 512$ Sentinel-1 SAR image pairs with simulated changes, testing model robustness against speckle noise and imaging geometries.
\end{itemize}

\begin{table}[t]
\small
\centering
\setlength{\tabcolsep}{4pt}
\begin{tabular}{lcccc}
\specialrule{1pt}{0pt}{0pt}
\textbf{Method} &
  \textbf{\begin{tabular}[c]{@{}c@{}}Params\\ (M)\end{tabular}} &
  \textbf{\begin{tabular}[c]{@{}c@{}}FLOPs\\ (G)\end{tabular}} &
  \textbf{IoU} &
  \textbf{F1} \\ \hline
FC-Siam-Diff \shortcite{daudt2018fully}         & 4.38  & 1.35   & 57.44 & 72.97 \\
FC-Siam-Conc \shortcite{daudt2018fully}         & 4.99  & 1.55   & 58.07 & 73.48 \\
STANet \shortcite{chen2020spatial}              & 16.93 & 6.58   & 65.66 & 79.31 \\
SNUNet \shortcite{fang2021snunet}               & 12.03 & 27.44  & 67.11 & 80.32 \\
BIT \shortcite{chen2021remote}                  & 3.55  & 4.35   & 70.64 & 82.80 \\
TransUNetCD \shortcite{li2022transunetcd}       & 28.37 & 244.54 & 71.86 & 83.63 \\
SwinSUNet \shortcite{zhang2022swinsunet}        & 39.28 & 43.50  & 74.82 & 85.60 \\
CTDFormer \shortcite{zhang2023relation}         & 3.85  & 303.77 & 67.09 & 80.30 \\
BiFA \shortcite{zhang2024bifa}                  & 5.58  & 53.00  & 72.35 & 83.96 \\
ConvFormer-CD/48 \shortcite{yang2025convformer} & 37.72 & 5.14   & 74.37 & 85.30 \\
SPMNet \shortcite{wang2025spmnet}               & 17.33 & 6.94   & 71.01 & 83.09 \\ \hline
\textbf{ChangeRWKV-T (ours)}                    & 4.66  & 9.40   & 75.14 & 85.80 \\
\textbf{ChangeRWKV-S (ours)}                    & 12.00 & 18.15  & 75.21 & 85.85 \\
\textbf{ChangeRWKV-B (ours)}                    & 20.50 & 33.56  & \textbf{75.46} & \textbf{86.01} \\
\specialrule{1pt}{0pt}{0pt}
\end{tabular}
\caption{Quantitative comparison on the LEVIR-CD+ \cite{shen2021s2looking} test set. All metrics are reported as a percentage (\%). Best results are highlighted in \textbf{bold}.}
\label{tab:levirp}
\end{table}

\noindent\textbf{Implementation Details.}
Following the setup of \cite{opencd}, all image pairs are cropped into fixed-size patches of $256 \times 256$ using a sliding window strategy. Our models are optimized with the Adam optimizer \cite{kingma2014adam}, using an initial learning rate of $10^{-5}$, a weight decay of $10^{-4}$, and a batch size of 8 per GPU. All experiments are conducted on four NVIDIA RTX 3090 GPUs. The learning rate follows a cosine annealing schedule, preceded by a warm-up phase of 20 epochs. Training proceeds for 200 epochs with a gradient clipping threshold of 0.5 to ensure numerical stability. To improve generalization, we adopt standard data augmentation techniques, including random rotation, flipping, brightness-contrast adjustment, and Gaussian blur, following FHD \cite{pei2022feature}.

To explore the trade-off between accuracy and efficiency, we instantiate ChangeRWKV in three model scales: ChangeRWKV-T (\textbf{T}iny), -S (\textbf{S}mall), and -B (\textbf{B}ase), which differ in embedding dimension and encoder depth. Specifically, their configurations are as follows:
\begin{itemize}
    \item \textbf{ChangeRWKV-T} uses embedding dimensions of [32, 48, 96, 160] and encoder depths of [2, 2, 4, 2].
    \item \textbf{ChangeRWKV-S} uses embedding dimensions of [32, 64, 128, 192] and encoder depths of [3, 3, 6, 3].
    \item \textbf{ChangeRWKV-B} uses embedding dimensions of [48, 72, 144, 240] and encoder depths of [3, 3, 6, 3].
\end{itemize}

\noindent\textbf{Evaluation Metrics}.
We use four standard metrics for quantitative evaluation: Precision (P), Recall (R), F1-score (F1), and Intersection over Union (IoU). IoU and F1 are the primary indicators of overall performance.

\begin{table}[t]
\small
\centering
\setlength{\tabcolsep}{4.8pt}
\begin{tabular}{lcccc}
\specialrule{1pt}{0pt}{0pt}
\textbf{Method} &
  \textbf{\begin{tabular}[c]{@{}c@{}}Params\\ (M)\end{tabular}} &
  \textbf{\begin{tabular}[c]{@{}c@{}}FLOPs\\ (G)\end{tabular}} &
  \textbf{IoU} &
  \textbf{F1} \\ \hline
FC-Siam-Diff \shortcite{daudt2018fully}           & 4.38  & 1.35   & 86.07 & 92.52 \\
FC-Siam-Conc \shortcite{daudt2018fully}           & 4.99  & 1.55   & 93.39 & 96.58 \\
BIT \shortcite{chen2021remote}                    & 3.55  & 4.35   & 93.48 & 96.63 \\
MFPNet \shortcite{xu2021remote}                   & 60.06 & 107.73 & 97.09 & 98.52 \\
MSCANet \shortcite{liu2022cnn}                    & 16.59 & 14.68  & 94.43 & 97.14 \\
FCS \shortcite{chen2022fccdn}                     & 3.03  & 9.59   & 94.77 & 97.31 \\
DED \shortcite{chen2022fccdn}                     & 3.10  & 9.90   & 94.94 & 97.40 \\
FCCDN \shortcite{chen2022fccdn}                   & 6.31  & 12.52  & 95.15 & 97.51 \\
ChangeFormer \shortcite{bandara2022transformer}   & 41.02 & 202.80 & 96.47 & 98.20 \\
AFCF-3D \shortcite{ye2023adjacent}                & 17.54 & 31.72  & 93.72 & 96.76 \\
ChangeMamba \shortcite{chen2024changemamba}       & 84.70 & 44.86  & 90.40 & 94.96 \\
\hline
\textbf{ChangeRWKV-T (ours)}                      & 4.66  & 9.40   & 96.87 & 98.41 \\
\textbf{ChangeRWKV-S (ours)}                      & 12.00 & 18.15  & 97.02 & 98.49 \\
\textbf{ChangeRWKV-B (ours)}                      & 20.50 & 33.56  & \textbf{97.18} & \textbf{98.57} \\
\specialrule{1pt}{0pt}{0pt}
\end{tabular}
\caption{Quantitative comparison on the SAR-CD \cite {alatalo2023improved} test set. All metrics are reported as a percentage (\%). Best results are highlighted in \textbf{bold}.}
\label{tab:sar}
\end{table}

\subsection{Comparison with State-of-the-Art Methods}
Benchmarked against leading state-of-the-art (SOTA) methods across four diverse datasets, ChangeRWKV sets a new standard in accuracy while establishing a superior trade-off between performance and computational cost.

\noindent\textbf{Performance on Standard Optical Benchmarks.}
As shown in Table \ref{tab:levir_whu}, our ChangeRWKV-B sets a new SOTA on the widely-used LEVIR-CD dataset, with an IoU of 85.46\% and an F1 of 92.16\%. This surpasses strong recent methods like ChangeBind \cite{noman2024changebind} (84.18\% IoU) and CBSASNet \cite{he2024cbsasnet} (84.14\% IoU). More importantly, our lightweight ChangeRWKV-T model, with just \textbf{4.66M} parameters and \textbf{9.40G} FLOPs, already outperforms most prior work with an IoU of \textbf{84.92\%}. On the WHU-CD dataset, our models maintain this strong performance, with the mid-sized ChangeRWKV-S achieving a top IoU of \textbf{90.06\%} and an F1 of \textbf{94.77\%}. These results confirm the effectiveness and scalability of the ChangeRWKV architecture on high-resolution optical imagery.

\noindent\textbf{Robustness to Long-Term Temporal Variations.}
The LEVIR-CD+ dataset presents a more challenging test with longer time intervals between images. As shown in Table \ref{tab:levirp}, ChangeRWKV excels in this setting. ChangeRWKV-B leads with a \textbf{75.46\%} IoU and an \textbf{86.01\%} F1, outperforming much heavier Transformer models like SwinSUNet (39.28M params) and ConvFormer-CD (37.72M params) while using only 20.5M parameters. The ability of our models to handle diverse, long-term changes underscores the effectiveness of our spatial-temporal fusion design and the representational power of the linear-attention backbone.

\noindent\textbf{Generalization to SAR Imagery.}
As detailed in Table \ref{tab:sar}, we evaluate the fundamental difference-learning capability of our models on SAR-CD, which features synthetic changes and significant speckle noise. Despite being designed for optical data, ChangeRWKV demonstrates remarkable generalization. The base model sets a new benchmark with an IoU of 97.18\% and an F1 of 98.57\%. Even the Tiny and Small variants surpass previous SOTA methods with a fraction of the parameters. This strong performance, without specific tuning for SAR data, highlights that our model learns a robust and modality-agnostic representation of change.

\begin{figure}[t]
\centering
\includegraphics[width=1.0\linewidth]{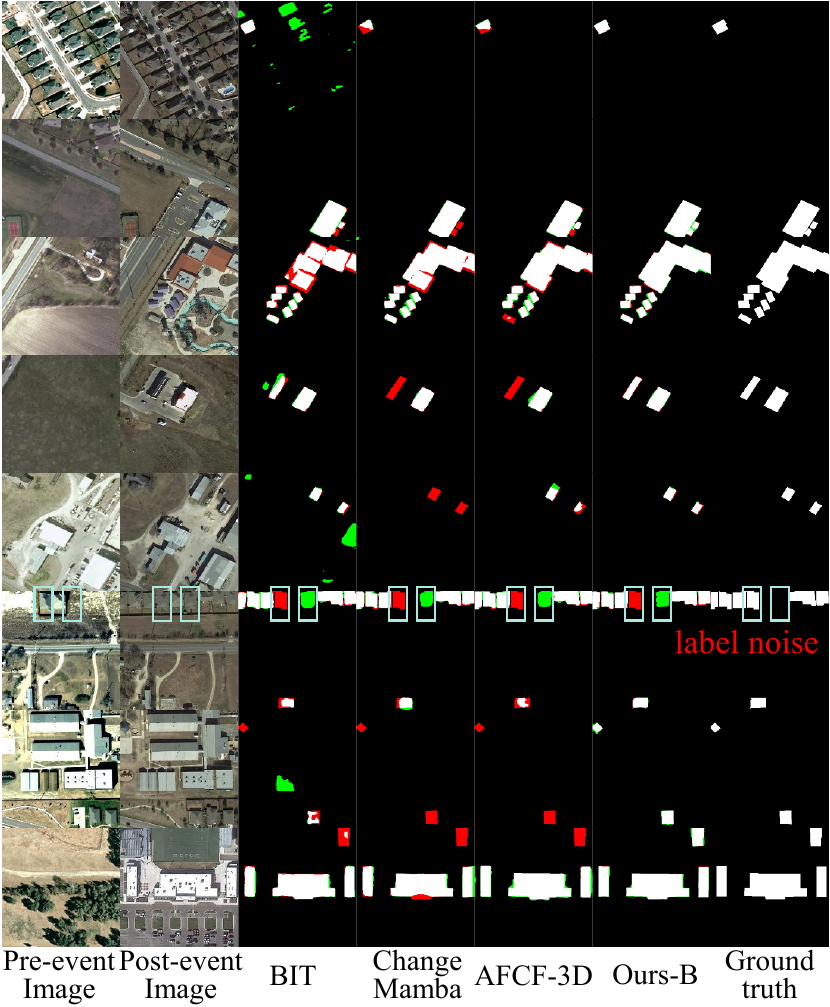}
\caption{Qualitative results on the LEVIR-CD test set. Predicted outputs are color-coded as follows: white for true positives (TP), black for true negatives (TN), green for false positives (FP), and red for false negatives (FN).}
\label{fig:vis}
\end{figure}

\begin{figure}[t]
\centering
\includegraphics[width=1.0\linewidth]{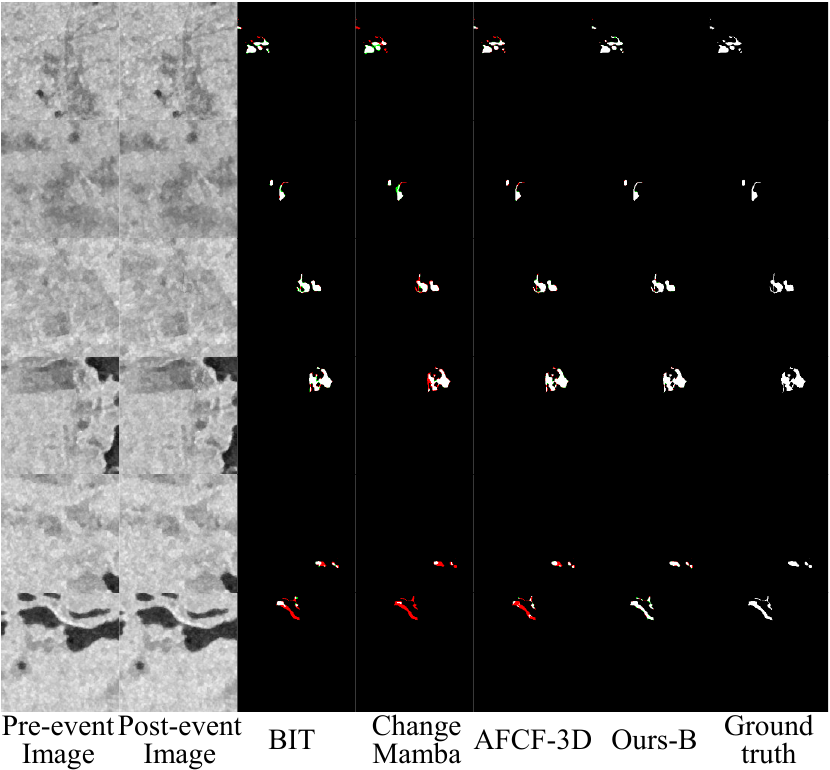}
\caption{Qualitative results on the SAR-CD test set. Predicted outputs are color-coded as follows: white for true positives (TP), black for true negatives (TN), green for false positives (FP), and red for false negatives (FN).}
\label{fig:sar}
\end{figure}

\noindent\textbf{Qualitative Analysis.}
As shown in Fig. \ref{fig:vis}, visual comparisons on LEVIR-CD confirm the superiority of our method. ChangeRWKV generates markedly sharper and more complete change masks, particularly in challenging scenarios where other methods struggle. It precisely delineates the boundaries of small, irregular buildings in dense urban layouts and maintains high fidelity for objects near image borders, a testament to the robustness of our hierarchical encoder and fusion module. Furthermore, our model effectively suppresses spurious predictions arising from background clutter and demonstrates resilience to common annotation noise \cite{sheng2024adaptive, sheng2024enhancing}, resulting in cleaner and more reliable change maps.
This robustness extends to different sensing modalities. Fig. \ref{fig:sar} illustrates this on SAR-CD, where ChangeRWKV produces clean and accurate detections despite significant speckle noise and non-semantic changes. This ability to perform well under heterogeneous conditions highlights the strong generalization of our framework, which learns fundamental patterns of change independent of the sensor type.

\begin{table}[t]
\small
\centering
\setlength{\tabcolsep}{4.6pt}
\begin{tabular}{ccccc}
\specialrule{1pt}{0pt}{0pt}
\textbf{Experiment}   & \textbf{Variant \#}    & \textbf{Method}         & \textbf{IoU}        & \textbf{F1} \\ \hline\hline
\textit{Component}    &                    &                     &            & \\ \hline
\multirow{3}{*}{\begin{tabular}[c]{@{}c@{}}Encoder\\ Architectures\end{tabular}}
    & \ding{182} & VisionRWKV                 & 81.86          & 90.03          \\
    & \ding{183} & + ViT-Adapter              & 83.06          & 90.75          \\
    & -          & \textbf{ours}              & \textbf{85.46} & \textbf{92.16} \\ \hline
\multirow{3}{*}{\begin{tabular}[c]{@{}c@{}}Channel \\ Mixing Module\end{tabular}}
    & \ding{184} & \textit{w/o}               & 84.43          & 91.56          \\
    & \ding{185} & RWKV-like                  & 85.38          & 92.11          \\
    & -          & \textbf{ours}              & \textbf{85.46} & \textbf{92.16} \\ \hline
\multirow{3}{*}{\begin{tabular}[c]{@{}c@{}}Spatial \\ Fusion Module\end{tabular}}
    & \ding{186} & \textit{w/o}               & 82.64          & 90.49          \\
    & \ding{187} & FPN                        & 83.56          & 91.05          \\
    & -          & \textbf{ours}& \textbf{85.46} & \textbf{92.16} \\ \hline
\multirow{5}{*}{\begin{tabular}[c]{@{}c@{}}Temporal \\ Fusion Module\end{tabular}}
    & \ding{188} & SiamDiff                   & 82.68          & 90.52          \\
    & \ding{189} & SiamConc                   & 82.73          & 90.55          \\
    & \ding{190} & FHD                        & 85.01          & 91.90          \\
    & \ding{191} & STRM                       & 83.13          & 90.79          \\
    & -          & \textbf{ours}              & \textbf{85.46} & \textbf{92.16} \\ \hline\hline
\textit{Hyperparameter}    &                  &                &              & \\ \hline
\multirow{4}{*}{\begin{tabular}[c]{@{}c@{}}Balance\\ Factor $\lambda$\end{tabular}}
    & \ding{172}      & 0                         & 84.95          & 91.86          \\
    & \ding{173}      & 0.5                       & 85.38          & 92.11          \\
    &  -              & \textbf{1.0 (ours)}          & \textbf{85.46} & \textbf{92.16} \\
    & \ding{174}      & 2.0                          & 85.12          & 91.96          \\
    \specialrule{1pt}{0pt}{0pt}
\end{tabular}
\caption{Ablation study on different architectural components. All metrics are reported as a percentage (\%).}
\label{tab:ablation}
\end{table}

\begin{figure*}[t]
    \centering
    \includegraphics[width=1\linewidth]{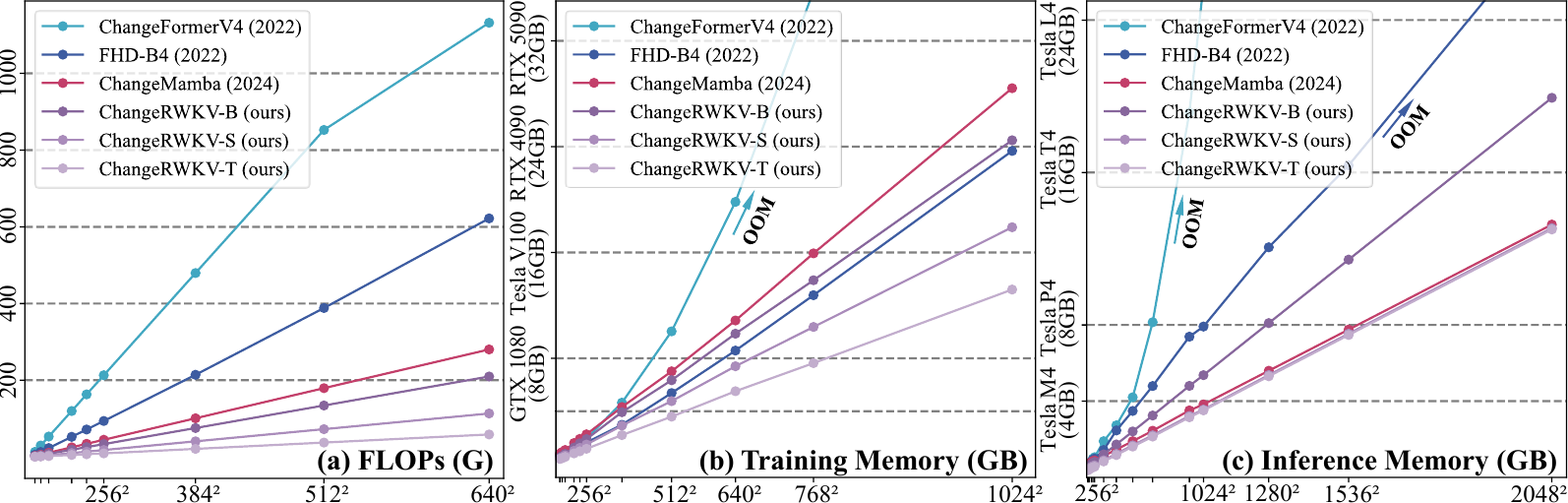}
    \caption{Computational Scalability Analysis. ChangeRWKV demonstrates near-linear growth in (a) FLOPs, (b) training memory, and (c) inference memory, significantly outperforming Transformer-based models at high resolutions.}
    \label{fig:effi}
\end{figure*}

\subsection{Ablation and Analysis}
We conduct a series of detailed experiments on LEVIR-CD to dissect the ChangeRWKV architecture and validate our design choices. We analyze the contribution of each core component and rigorously evaluate the model's computational efficiency, with results summarized in Table \ref{tab:ablation}.

\noindent\textbf{Encoder Architecture (\textit{cf.} \ding{182} \ding{183}).}
To validate our hierarchical design, we compare it against two baselines: a ``flat'' VisionRWKV encoder \cite{duan2024vision} and the original VisionRWKV design which uses a ViT-Adapter \cite{chen2022vitadapter} for segmentation. The results show a clear progression: simply adding the adapter for hierarchical features improves the IoU from 81.86\% to \textbf{83.06\%}. Our tailored hierarchical encoder further boosts performance to \textbf{85.46\%} IoU and \textbf{92.16\%} F1, confirming that a specialized multi-scale architecture is essential for effective change detection.

\noindent\textbf{Channel Mixing Module (\textit{cf.} \ding{184} \ding{185}).}
We analyze the trade-off between performance and computational cost in our channel mixing design. Compared to our lightweight SE-based module, a heavier, standard RWKV-style mixing block provides a negligible performance difference (85.38\% vs. \textbf{85.46\%} IoU). However, our design is vastly more efficient, reducing computational cost from 71.66G to \textbf{33.56G} FLOPs and parameters from 47.1M to \textbf{20.5M}. This demonstrates that our lightweight module effectively captures feature interactions at less than half the computational budget.

\noindent\textbf{Spatial Fusion Module (\textit{cf.} \ding{186} \ding{187}).}
The SFM is designed to align features across scales before temporal comparison. Removing SFM or replacing it with an FPN-style \cite{lin2017feature} fusion degrades IoU by \textbf{2.82\%} and \textbf{1.90\%}, respectively, highlighting that enforcing intra-image spatial consistency is critical for precise change localization.

\noindent\textbf{Temporal Fusion Module (\textit{cf.} \ding{188} \ding{189} \ding{190} \ding{191}).}
We benchmark our cross-attention based TFM against several common fusion strategies. Our module significantly outperforms simpler methods like SiamDiff (82.68\% IoU) and SiamConc (82.73\% IoU), as well as more recent techniques like FHD \cite{pei2022feature} (85.01\% IoU) and STRM \cite{chen2024changemamba} (83.13\% IoU). The superior performance of our TFM highlights its advanced capability to model complex, fine-grained temporal interactions between feature pairs.

\noindent\textbf{Loss Function Balance (\textit{cf.} \ding{172} \ding{173} \ding{174}).}
We finally examine the weighting factor $\lambda$ for the Dice loss component. We found that a balanced contribution ($\lambda=1.0$) provides the best results, confirming that a joint objective optimizing both pixel-level accuracy (BCE) and region-level coherence (Dice) is optimal for the remote sensing change detection task.

\subsection{Efficiency and Scalability Analysis}
A core motivation for ChangeRWKV is to address the computational bottleneck in high-resolution change detection. We evaluate our models' resource consumption for inputs ranging from $64^2$ to $2048^2$ pixels. As shown in Fig. \ref{fig:effi}, ChangeRWKV's linear-time attention mechanism provides a clear advantage. While Transformer-based methods exhibit quadratic growth, our models' resource usage scales near-linearly.
This efficiency is stark in direct comparison: at a $512 \times 512$ resolution, our ChangeRWKV-B requires only \textbf{134.25G} FLOPs and\textbf{ 6.5GB} of training memory, a sharp contrast to the 852.44G FLOPs and 10.2GB required by ChangeFormer. This makes our approach vastly more scalable. Critically, this efficiency translates to real-world deployability. All three ChangeRWKV variants can perform inference on $1024^2$ inputs using a single NVIDIA Tesla P4 GPU with only 8GB of VRAM, highlighting its suitability for resource-constrained and edge-computing scenarios.

\section{Conclusion}
This paper presents ChangeRWKV, an \textit{efficient} and \textit{powerful} framework for remote sensing change detection that directly addresses the trade-off between accuracy and computational cost. Our approach leverages a hierarchical encoder built upon the \textit{linear-time} RWKV architecture to capture multi-scale spatial features with exceptional efficiency. The core of our framework is the novel Spatial-Temporal Fusion Module (STFM), which adaptively integrates bi-temporal features to highlight meaningful changes. Comprehensive experiments show that ChangeRWKV establishes a new state-of-the-art on multiple optical and SAR benchmarks, delivering superior performance while requiring only a fraction of the computational resources of previous methods.

\noindent\textbf{Limitations and Future Work.}
Despite its strong performance, our work's limitations include the reliance on large annotated datasets and a lack of modality-specific priors (\textit{e.g.}, for SAR or hyperspectral data). Future work will focus on exploring weakly-supervised learning and enabling real-time deployment on resource-constrained platforms.

\section{Acknowledgments}
This work was supported by the National Defense Science and Technology Industry Bureau Technology Infrastructure Project (JSZL2024606C001), Key Research and Development (R\&D) Plan Project of Jiangsu Province (No. BE2023008-2).

\small\bibliography{aaai2026}

\end{document}